\documentclass[letterpaper]{article} 
\usepackage{aaai25}  
\usepackage{times}  
\usepackage{helvet}  
\usepackage{courier}  
\usepackage[hyphens]{url}  
\usepackage{graphicx} 
\usepackage{natbib}  
\usepackage{caption} 
\usepackage{algorithm}
\usepackage{algorithmic}
\usepackage{array}
\usepackage{booktabs}
\usepackage{newfloat}
\usepackage{listings}
\DeclareCaptionStyle{ruled}{labelfont=normalfont,labelsep=colon,strut=off} 
\floatstyle{ruled}
\newfloat{listing}{tb}{lst}{}
\floatname{listing}{Listing}
\title{Do Students Rely on AI? Analysis of Student-ChatGPT Conversations from a Field Study}
\author {
    Jiayu Zheng\textsuperscript{\rm 1}, 
    Lingxin Hao\textsuperscript{\rm 1}, 
    Kelun Lu\textsuperscript{\rm 2}, 
    Ashi Garg\textsuperscript{\rm 1}, 
    Mike Reese\textsuperscript{\rm 1}, 
    Melo-Jean Yap\textsuperscript{\rm 1}, 
    I-Jeng Wang\textsuperscript{\rm 3}, 
    Xingyun Wu\textsuperscript{\rm 1}, 
    Wenrui Huang\textsuperscript{\rm  1}, 
    Jenna Hoffman\textsuperscript{\rm 1}, 
    Ariane Kelly\textsuperscript{\rm 1}, 
    My Le\textsuperscript{\rm 1}, 
    Ryan Zhang\textsuperscript{\rm 1}, 
    Yanyu Lin\textsuperscript{\rm 1}, 
    Muhammad Faayez\textsuperscript{\rm 1}, 
    Anqi Liu\textsuperscript{\rm 1}
}
\affiliations {
    \textsuperscript{\rm 1} Johns Hopkins University, Baltimore, MD, USA,
    
    \textsuperscript{\rm 2} University of Pennsylvania, Philadelphia, PA, USA,

    \textsuperscript{\rm 3} Johns Hopkins University Applied Physics Lab, Laurel, MD, USA

    jzheng59@jh.edu, hao@jhu.edu, 
    aliu74@jhu.edu
}

\begin{document}

\maketitle

\begin{abstract}
This study explores how college students interact with generative AI (\texttt{ChatGPT-4}) during educational quizzes, focusing on reliance and predictors of AI adoption. Conducted at the early stages of \texttt{ChatGPT} implementation, when students had limited familiarity with the tool, this field study analyzed 315 student-AI conversations during a brief, quiz-based scenario across various STEM courses. A novel four-stage reliance taxonomy was introduced to capture students' reliance patterns, distinguishing AI competence, relevance, adoption, and students' final answer correctness. Three findings emerged. First, students exhibited overall low reliance on AI and many of them could not effectively use AI for learning. Second, negative reliance patterns often persisted across interactions, highlighting students’ difficulty in effectively shifting strategies after unsuccessful initial experiences. Third, certain behavioral metrics strongly predicted AI reliance, highlighting potential behavioral mechanisms to explain AI adoption. The study's findings underline critical implications for ethical AI integration in education and the broader field. It emphasizes the need for enhanced onboarding processes to improve student's familiarity and effective use of AI tools. Furthermore, AI interfaces should be designed with reliance-calibration mechanisms to enhance appropriate reliance. Ultimately, this research advances understanding of AI reliance dynamics, providing foundational insights for ethically sound and cognitively enriching AI practices.
\end{abstract}

\section{Introduction}
The rapid advancement and pervasive integration of artificial intelligence (AI) systems are profoundly reshaping diverse domains, with education emerging as a significant frontier \cite{costa2024critical, wu2025systematic, hasanein2023drivers, firat2025global}. AI holds immense potential to revolutionize educational practices by offering tailored guidance, feedback, and support through personalized learning platforms and intelligent tutoring systems \cite{wang2024artificial, paladines2020systematic, galdames2024impact}. These technologies promise to enhance student engagement, improve academic outcomes, and adapt instruction to individual learning patterns \cite{wang2024artificial}. However, the increasing integration of AI also necessitates a critical examination of how humans interact with and rely on these systems, which will inform ethical deployment of AI in education applications.

One critical concern is that students might over-rely on AI, which influences critical thinking \cite{gerlich2025ai, zhai2024effects}, human decision-making, analytical thinking \cite{grassini2023shaping} and increases human laziness \cite{ahmad2023impact}. Previous studies have extensively explored specific aspects of AI reliance, typically within lab settings or field experiments \cite{dai2025generative}. However, some areas of AI reliance are underexplored. First, the understanding of spontaneous human-AI interactions, how students naturally utilize AI tools in authentic academic contexts \cite{eckhardt2024survey} and the verification of real-world data, particularly concerning the underlying behavioral mechanisms that explain different reliance scenarios remains limited \cite{dai2025generative}. Furthermore, existing research predominantly takes a static view to lable messages or conversations without capturing the evolving nature of the student-AI conversation dynamically \cite{solomon2022using}. Lastly, most reliance definitions in the human–AI interaction and AI-assisted decision-making literature rely on initial human judgments and assume that they can be labeled simply as correct or incorrect. This binary framing does not translate well to educational settings, where it is hard to evaluate students' initial status and students are actively acquiring new knowledge with AI assistance.

This study addresses these gaps by conducting an examination of students’ reliance on AI within an authentic university-level STEM quiz environment using a novel four-stage reliance taxonomy in 315 conversations. Through comprehensive behavioral and text analyses, we identify patterns of student reliance and elucidate underlying factors influencing reliance behaviors. In a college peer-led-team-learning (PLTL) program \cite{bauer2025students}, students place greater trust in the instructors, teaching assistants, and knowledgeable peers than in GenAI given the inherent property of uncertainty in AI responses and the unfamiliarity of college students with \texttt{ChatGPT} when it was first released. Our field study took advantage of this real-world setting by deploying an interface with \texttt{ChatGPT-4} and a course-specific quiz application for 30 minutes during a PLTL session of STEM gateway courses\footnote{More details about the PLTL program can be found at \url{https://academicsupport.jhu.edu/pilot/}.}. Of the consented students, 182 students decided to use the interface for solving at least one quiz problem. The study was approved by the university IRB. The results will shed light on an understanding of the trust in \texttt{ChatGPT} when PLTL program is available. In such a context, we believe that the degree to which students' engagement with \texttt{ChatGPT} and the ways by which students become critically reliant on \texttt{ChatGPT} will reveal contributing factors of reliance and the underlying ethical issues.

Our study makes several detailed contributions. Firstly, we develop a reliance taxonomy that extends beyond traditional reliance metrics by incorporating measures of AI correctness, AI relevance, students' adoption of AI advice and the correctness of final answer. Secondly, we provide both conversation-level and individual-level analyses, thereby capturing reliance dynamics over time and highlighting distinct reliance trajectories that students exhibit across multiple interactions. Finally, we utilize behavioral and text features to identify predictors of different types of AI adoption, uncovering crucial mechanisms that govern students' AI adoption patterns and offering insights for enhancing human-AI interaction design.

\section{Related Work}
\subsection{Human-AI Interaction in Education}
AI has been integrated into various areas of education, offering diverse applications that influence students' academic development. These applications include intelligent tutoring systems, educational robots, learning analytics dashboards, and adaptive learning platforms \cite{galdames2024impact, wang2024artificial}. Generative AI (GenAI) tools, such as \texttt{ChatGPT}, are increasingly utilized in higher education, providing personalized learning and assessment opportunities \cite{wu2025systematic}. They have the potential to adapt instruction to different student types, offer feedback, assist in course design, and support academic writing \cite{santos2024generative}.

The influence of AI on students, however, presents inconsistent findings in the literature. Some studies highlight the benefits, suggesting that AI can enhance learning outcomes, improve knowledge accessibility, promote personalized support, and increase student motivation and engagement \cite{ding2023students, ouyang2021artificial, singh2023exploring}. For instance, AI-powered platforms have been shown to enhance student engagement and performance through real-time feedback and customized instructions \cite{hakiki2023exploring}. Conversely, other research points to potential drawbacks. Concerns include the risk of over-reliance on AI, which may hinder the development of critical thinking and independent problem-solving skills \cite{kosar2024computer, krupp2024unreflected}. This often manifests by students reducing their mental effort and relying on AI for quick solutions, bypassing deeper cognitive processes essential for learning \cite{fan2025beware}. Ethical issues are also prominent, particularly concerning academic dishonesty and the uncritical acceptance of AI-generated content \cite{acosta2024knowledge, tan2023impact}. The quality of AI-generated content can be inconsistent, inaccurate, or lack depth, depending on user prompts, potentially leading to misunderstandings or the absorption of misleading information \cite{wang2023survey}.

A significant gap in current research is that many studies employ experimental designs to investigate how specific factors influence human-AI interaction in controlled educational settings \cite{dai2025generative}. While valuable, this approach often fails to capture how humans use ``natural AI in the wild''—that is, how students spontaneously interact with widely available AI tools like \texttt{ChatGPT} in their everyday learning processes. Understanding these natural usage patterns from the user's perspective is crucial for inferring how AI systems are truly adapted and integrated into learning behaviors, providing insights beyond what can be observed in structured experiments.
\subsubsection{Behavioral analysis of human-AI interaction in education}
Understanding human-AI interactions necessitates a granular analysis of user behavior while little attention has been paid to this matter. In educational settings, behavioral analysis provides a powerful lens for revealing patterns and trends in learning behaviors, which can inform the design and improvement of AI-supported educational platforms \cite{dai2025generative}. 

Scholars have used several behavioral factors to understand student-AI interactions. Interaction patterns and sequences are meticulously studied, revealing how students engage with AI systems through specific behavioral flows, and the routines of human behaviors \cite{sun2024human}, such as querying background information, periods of idle operation, reviewing project requirements, and interpreting AI-generated content \cite{dai2025generative}. A study also categorizes students’ usage of AI by the combination of ``mode of interaction'' and ``desired outcome'' as well as using Bloom's Taxonomy to classify cognitive processes from simpler to more complex \cite{handa2025education} . As suggested by their findings, students delegate higher-order cognitive tasks to AI systems. How do we ensure students still have the capabilities to discern the correctness of AI’s answers and do not over-rely on AI?

Despite these advancements, a significant gap in behavioral analysis of student-AI interaction lies in the focus on message-level analysis rather than considering the conversation as a whole unit. Many current research efforts tend to analyze individual turns or isolated messages, potentially overlooking the holistic, evolving nature of student-AI interaction over a sustained period. Some research captures behaviors and conversations in human-AI interaction \cite{mcnichols2025studychat, dai2025generative, ammari2025students}. However, capturing the full context and flow of a conversation, including how students iteratively refine their questions or respond to prior AI turns, is crucial for a comprehensive understanding of reliance dynamics. This necessitates approaches that can analyze the entire student-AI conversation as a cohesive unit, rather than just discrete messages.

\subsection{Reliance Analysis in Human-AI Interaction}
\subsubsection{Reliance in the literature}
In the literature, reliance is defined as the observable behavior of a user following or utilizing AI advice \cite{eckhardt2024survey}. However, relying on AI's wrong advice would lead to incorrect decisions. Therefore, scholars also introduce ``appropriateness'' in reliance analysis \cite{schemmer2023appropriate}.

Various metrics have been employed to quantify AI reliance, often depending on the decision-making context. The ``agreement percentage'' quantifies how often a user's decision matches the AI's advice. The ``switch percentage'' measures how often users change their initial decision to align with AI advice, typically in a two-stage decision process where an initial human decision is recorded before AI advice is presented. The ``Weight of Advice (WOA)'' quantifies the impact of AI recommendations on a user's final decision, particularly in non-discrete scenarios \cite{eckhardt2024survey}. Appropriateness also measures the percentage in which the decision-maker relies on correct AI advice and does not rely on incorrect AI advice.

To further distinguish whether the reliance on correct AI advice stems from a correct discrimination or simply an overlap of the human and the AI's decision, judge-advisor system paradigm includes an initial human decision, correctness of AI's advice and whether human follow is adopted \cite {schemmer2023appropriate, eckhardt2024survey, cao2024designing}. However, in the real-world educational scenarios, particularly when students are learning, they may not have an initial decision or a pre-existing answer to a problem. Instead, they might consult AI as a primary source of information or a tool for exploration, making the concept of an ``initial decision'' irrelevant. This highlights the need for a reliance framework that does not rely on unobservable initial human decisions, allowing for a more accurate assessment of reliance in dynamic learning contexts.

To achieve the potential of human-AI collaboration with a better performance than either the human or the AI alone, researchers advocate the need the human to exercise discretion in following AI's advice, i.e., appropriately relying on the AI’s advice \cite{schemmer2023appropriate}. Recognizing the importance of building a mental model of the AI to assess AI recommendations, i.e., humans having an accurate understanding of what the AI system can and cannot do, researchers hypothesize and demonstrate that human learning is a key mediator of appropriate reliance in an experiment with 100 participants \cite{schemmer2023appropriate}. 

\subsubsection{What factors influence reliance?}
Human's reliance on AI systems is a multifaceted phenomenon influenced by a complex interplay of system-related, user-related, and task-related factors \cite{eckhardt2024survey}.

The intrinsic characteristics of the AI system itself are paramount in shaping user reliance, such as AI’s accuracy \cite{lai2019human, yin2019understanding, kahr2024understanding}, explainability \cite{ bansal2021does, kahr2024understanding}, alignment with human’s values \cite{narayanan2023does}, and reliability of AI \cite{zhai2024effects}. Ethical issues of AI, including hallucination, algorithmic bias, overconfidence, plagiarism, privacy concerns and tranparency concerns influence people's trust in and adoption of AI. Besides intrinsic problems with the system, how human interacts with AI also affects human-AI reliance. For example, ``helpfulness'' is an important aspect we need to take into consideration in the reliance framework, as AI might not understand users’ requests when users do not provide accurate prompts to reflect their needs, which hinders human from appropriately relying on AI.

Intrinsic individual user characteristics will influence reliance. A higher general inclination to trust technology can lead to greater reliance on correct AI advice \cite{li2023modeling, kahr2024understanding, kuper2025psychological}. Users with higher domain expertise tend to exhibit higher openness to technology yet lower levels of trust in and reliance on AI systems \cite{ kuper2025psychological}. A user's confidence in their initial decision and prior experience can shape their susceptibility to AI recommendations, with high confidence correlating with greater self-reliance, particularly when the initial decision is correct \cite{ schemmer2023appropriate, zhou2024rel,cao2024designing}. 

The nature of the task also influences reliance. User reliance can vary based on the task type (e.g., objective vs. subjective) and its inherent complexity \cite{ schaschek2024those}.

\subsubsection{Reliance in education applications} Student's reliance on AI in education is a growing area of concern, with mixed findings regarding its implications. Although utilizing \texttt{ChatGPT} has the advantage of increasing efficiency by saving time and effort, users could get into the habit of adopting the answers without rationalization or verification. Over-reliance on GenAI also contributes to negative outcomes such as influencing critical thinking \cite{gerlich2025ai, zhai2024effects}, human decision-making, analytical thinking \cite{grassini2023shaping} and increasing human laziness \cite{ahmad2023impact}. This evidence underscores the complexity of integrating such powerful tools into learning environments, especially when students may simply copy and paste AI's answers without critical evaluation, undermining academic integrity and hindering their own learning.

However, there is a notable lack of research specifically on human's reliance on AI in education with verification of real-world data, particularly concerning the underlying mechanisms that explain different reliance scenarios \cite{dai2025generative}. 
While some studies analyze reliance by examining how students use AI for specific tasks or how their performance changes with AI assistance, there is a significant gap in using behavioral analysis to explain why different reliance scenarios occur. Existing research often describes reliance outcomes but lacks a deeper investigation into the interactive mechanisms and features that drive these behaviors. This absence of behavioral analysis limits our understanding of the complex interplay between student actions, AI responses, and the resulting reliance patterns in educational settings.

\subsection{This Study}
The preceding review highlights several research gaps in the understanding of human-AI reliance, particularly within educational contexts. First, there is a lack of research on the user's perspective of human-AI interaction when users engage with ``natural AI in the wild'', rather than in controlled experimental settings \cite{eckhardt2024survey}. This limits our understanding of how students spontaneously integrate and adapt to AI tools in their authentic learning environments. Second, despite the growing sophistication of behavioral analysis, there is a continuing need for an approach that can capture the dynamic patterns that predict reliance, especially in complex, open-ended conversational settings. Specifically, much of the current research focuses on message-level analysis, rather than considering the conversation as a whole unit, which limits the ability to understand the holistic, evolving nature of student-AI interaction over time \cite{solomon2022using}. Third, there is a general lack of research on human reliance on AI in education, especially concerning the underlying mechanisms that explain different reliance scenarios. Current reliance frameworks often rely on the presence of an initial human decision, which may not be applicable in learning contexts where students are actively acquiring new knowledge with AI assistance.

Our paper attempts to fill this substantial gap using the analysis of 315 factual student-AI conversations from a field study. In this analysis of students' dialogues with \texttt{ChatGPT}, we address the ethical quandaries posed by GenAI within an authentic higher education setting. Our paper attempts to provide a nuanced understanding of how students are engaged in exchanging with \texttt{ChatGPT} for the learning task around a quiz problem. This design advances our understanding of whether and how \texttt{ChatGPT} influences students’ cognitive skills in solving problems through  Our ultimate goal is that the use of GenAI to assist college students' learning is ethically sound and cognitively enriching. This research seeks to answer the following questions.

\begin{enumerate}
    \item How do students rely on AI when using AI for a quiz?
    \item What factors are related to students' adoption of AI's advice?
\end{enumerate}

\section{Data}
We conducted a field study within the PLTL program at a university in weeks 9-11, Fall 2023. The study leveraged a web-based quiz application that integrated \texttt{ChatGPT-4} as an AI learning peer. In the first eight weeks, students had regular PLTL interactions with human learning peers. During the first 30 mininutes of the PLTL sessions in weeks 9-11, human learning peers were replaced with the AI learning peer. Among the consented students, 216 students were assigned to use AI but only 182 of them interacted with \texttt{ChatGPT-4} for at least one quiz problem. The time-stampted data include quiz questions and students' answers, and logs of student–AI conversations. The unit of analysis is the conversation, defined as the interaction between a student and \texttt{ChatGPT-4} for a quiz problem. Focusing on five subject areas (mathematics, physics, chemistry, economics, and public health), the analytic data comprises 315 student–AI conversations from 182 students with valid textual contents. The quiz answers were graded by the teaching assistants according to the grading instructions provided by the instructors.

\section{Measurements}
\subsection{Reliance Taxonomy} We develop a four‐stage reliance taxonomy to capture students’ interaction with \texttt{ChatGPT} during quiz tasks. The four stages are: (1) \texttt{ChatGPT’s} competence (factual correctness), (2) \texttt{ChatGPT’s} relevance (whether it addresses the student's question), (3) student's adoption of \texttt{ChatGPT’s} advice, and (4) correctness of the student's final answer. First, we determine whether \texttt{ChatGPT’s} response is factually correct. Next, we assess whether a correct response actually addresses the student’s question (an incorrect response cannot be relevant by definition). Third, we code whether the student follows \texttt{ChatGPT’s} suggested solution path. Finally, we evaluate whether the student’s submitted answer is correct.  

\begin{figure*}[tb]
  \centering
  \includegraphics[width=0.95\textwidth]{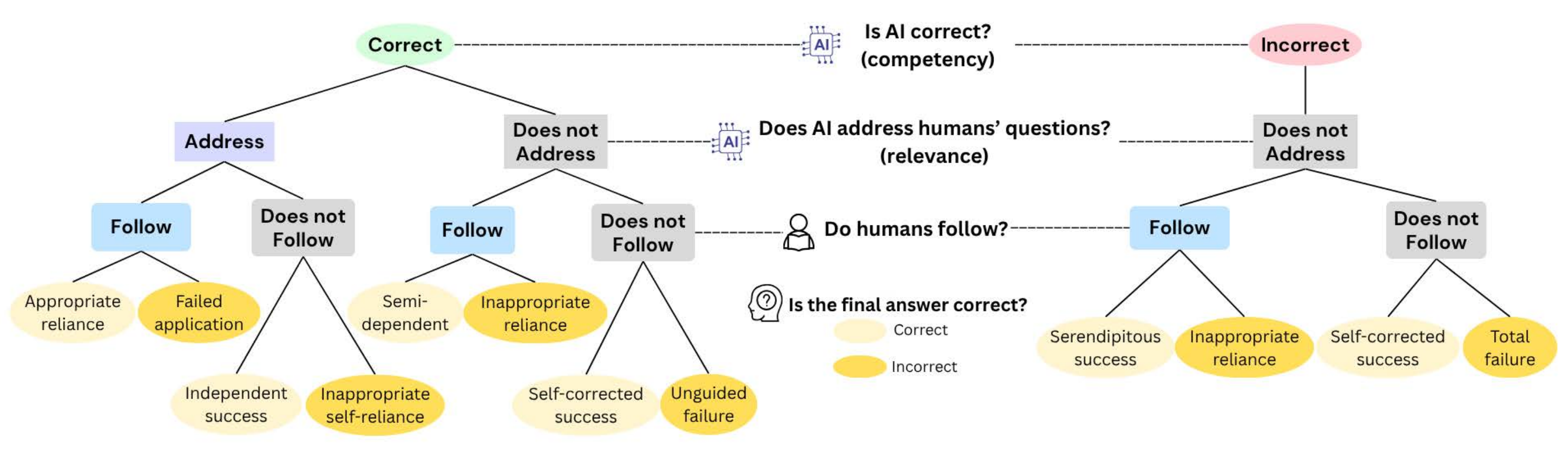}
  \caption{Visualizing Reliance Taxonomy. Based on our identification of the 4 labels as the \textit{reliance code}, we classify student-AI conversation into 12 scenarios. If the answer given by AI is factually wrong at the first place, the answer is irrelevant to student's question by definition. Link to the figure: \url{https://www.canva.com/design/DAGn8nxHMCg/O0aiU3cLlVlY15fD8HcKpA/edit}.}
  \label{fig:correct1}
\end{figure*}

Concatenating these four binary labels produces a four‐digit \emph{reliance code} (e.g. 1–1–1–1 indicates that \texttt{ChatGPT's} answer is correct (1), and relevant (1), the student followed (1), and the final answer is correct (1); 0-0-0-0 indicates that \texttt{ChatGPT's} answer is incorrect (0), and irrelevant (0), the student did not follow (0), and the final answer is incorrect (0)), which we then map to one of twelve \emph{reliance scenarios} (e.g.\ “Appropriate reliance,” “Failed application,” “Independent success,” “Inappropriate self‐reliance,” “Semi‐dependent,” “Self‐corrected success,” “Unguided failure,” “Serendipitous success,” and “Total failure”). We define "Appropriate reliance" as students leveraging AI to obtain help that supports arriving at a correct answer but not considering cheating i.e., copy and paste quiz question text for quick answer. To verify this, we compute similarity scores between each student’s prompt and the corresponding quiz question text. The observed average similarity is low, suggesting that students typically paraphrased the question or asked their own variants instead of submitting the original text verbatim, which reduces concern about simple shortcutting or cheating. Table~\ref{tab:reliance_taxonomy} provides a full description of each category. We also visualize the taxonomy for the classification in Figure~\ref{fig:correct1}.

\begin{table*}[tb]
\centering
\setlength{\tabcolsep}{2pt}  
\small                
\begin{tabular}{
    p{0.06\textwidth}   
    p{0.07\textwidth}   
    p{0.07\textwidth}   
    p{0.07\textwidth}   
    p{0.07\textwidth}   
    p{0.16\textwidth}   
    p{0.45\textwidth} 
@{}}
\hline
GPT correct? & GPT relevant? & Student follow? & Final ans.\ correct? 
  & Reliance code & Reliance label & Interpretation \\
\hline
1 & 1 & 1 & 1 & 1-1-1-1 & Appropriate reliance        
  & GPT gave a correct, relevant answer, student followed it, and got the right result. \\

1 & 1 & 1 & 0 & 1-1-1-0 & Failed application          
  & GPT guidance was correct and relevant, student followed, but the final outcome was wrong. \\

1 & 1 & 0 & 1 & 1-1-0-1 & Independent success         
  & GPT was correct and relevant, student did not use it but solved the problem. \\

1 & 1 & 0 & 0 & 1-1-0-0 & Inappropriate self-reliance 
  & GPT was accurate and relevant, yet student ignored it and failed. \\

1 & 0 & 1 & 1 & 1-0-1-1 & Semi-dependent              
  & GPT answer was correct but off-topic; student followed nonetheless got it right. \\

1 & 0 & 1 & 0 & 1-0-1-0 & Inappropriate reliance       
  & GPT was correct but irrelevant, student followed it and failed. \\

1 & 0 & 0 & 1 & 1-0-0-1 & Self-corrected success       
  & GPT was correct but irrelevant; student ignored it and succeeded. \\

1 & 0 & 0 & 0 & 1-0-0-0 & Unguided failure            
  & GPT was correct yet irrelevant; student ignored it and failed. \\

0 & 0 & 1 & 1 & 0-0-1-1 & Serendipitous success       
  & GPT gave a wrong and off-topic answer; student followed it but arrived at the correct result. \\

0 & 0 & 1 & 0 & 0-0-1-0 & Inappropriate reliance      
  & GPT was wrong and irrelevant; student followed it and failed. \\

0 & 0 & 0 & 1 & 0-0-0-1 & Self-corrected success       
  & GPT was wrong and irrelevant; student ignored it and succeeded. \\

0 & 0 & 0 & 0 & 0-0-0-0 & Total failure               
  & GPT was wrong and irrelevant; student ignored it and failed. \\
\hline
\end{tabular}
\caption{Reliance Taxonomy of Human--AI Reliance Scenarios}
\label{tab:reliance_taxonomy}
\end{table*}

\subsection{Behavioral and Text Features Extraction} From the student-GPT conversations, we generated behavioral features and text features. 
\subsubsection{Behavioral features} describe how students interact with the chatbot. Specifically, interaction time in seconds (mean: 180.26, standard deviation (SD): 278.55) captures the total duration from the user’s initial prompt to the bot’s final response, while answer time in seconds (mean: 122.55, SD: 274.67) measures the interval from the bot’s last message to the submission of an answer. Generally, students spent more time interacting with AI than answering the quiz problems. The average interaction time is around three minutes.

\begin{figure}[tb]
  \centering
  \includegraphics[width=0.9\columnwidth]{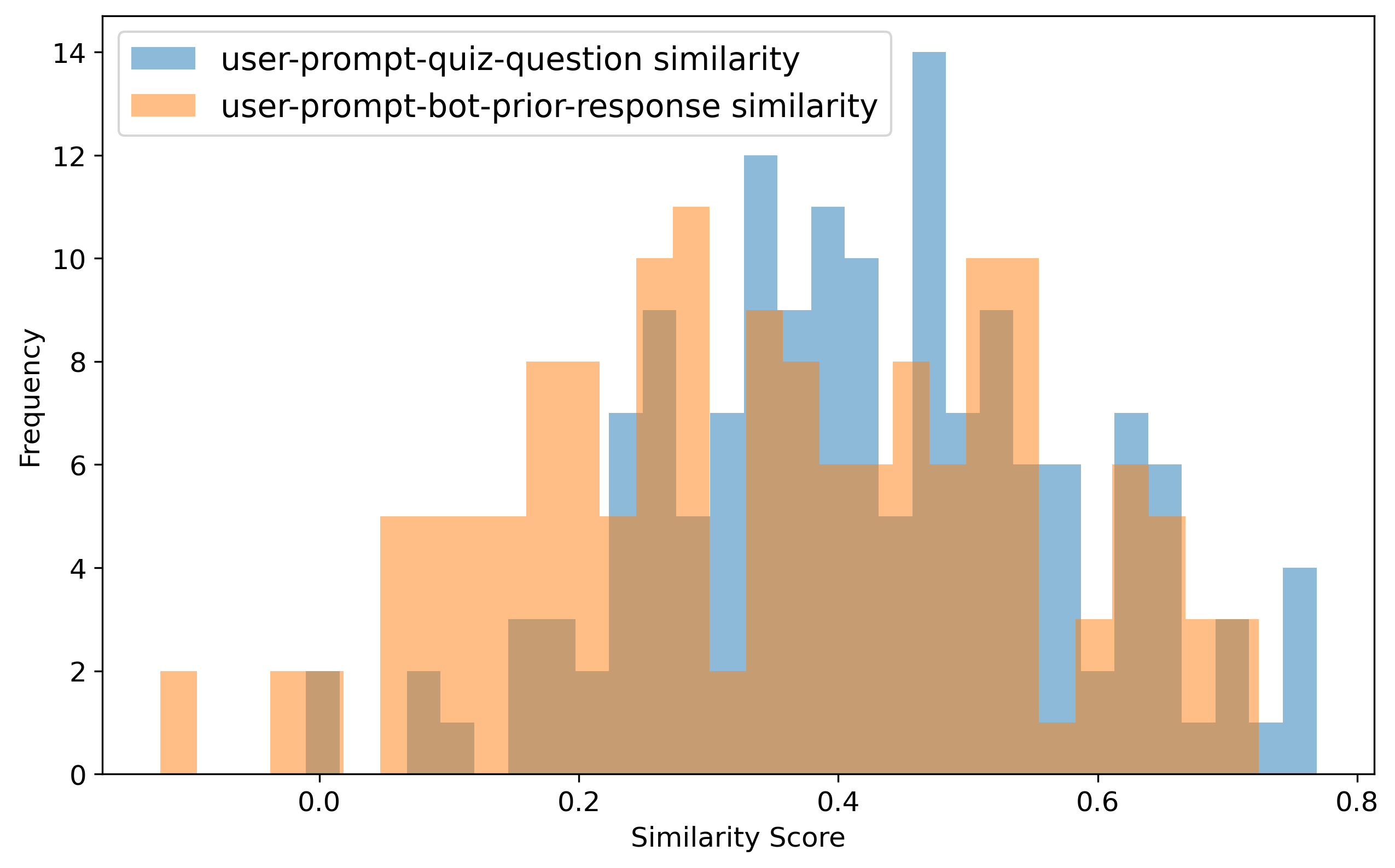}
  \caption{Distribution of Similarity Scores}
  \label{fig:similarity}
\end{figure}

Two similarity metrics are used to measure how students make use of AI to answer the quiz problems. First, user‑prompt‑quiz‑question similarity (mean: 0.50, SD: 0.23) assesses the extent to which users copy the quiz question text or rephrase it. Second, user‑prompt‑bot‑prior‑response similarity (mean: 0.17, SD: 0.22) evaluates to what extent students' prompts are related to the bot’s previous response. Both simialrity metrics are measured by \texttt{all-distilroberta-v1}, which is intented to be used as a sentence and short paragraph encoder, fitting with our quiz context. In general, students did not simply copy and paste the text of the quiz questions to get a quick answer. Instead, they paraphrased the questions or asked their own questions. The low user‑prompt‑bot‑prior‑response similarity suggests that students' prompts are not entirely bonded with AI's response. The distributions of simialrity scores of both users' and bot's messages are shown in Figure \ref{fig:similarity}.

\begin{figure}[tb]
  \centering
  \includegraphics[width=0.9\columnwidth]{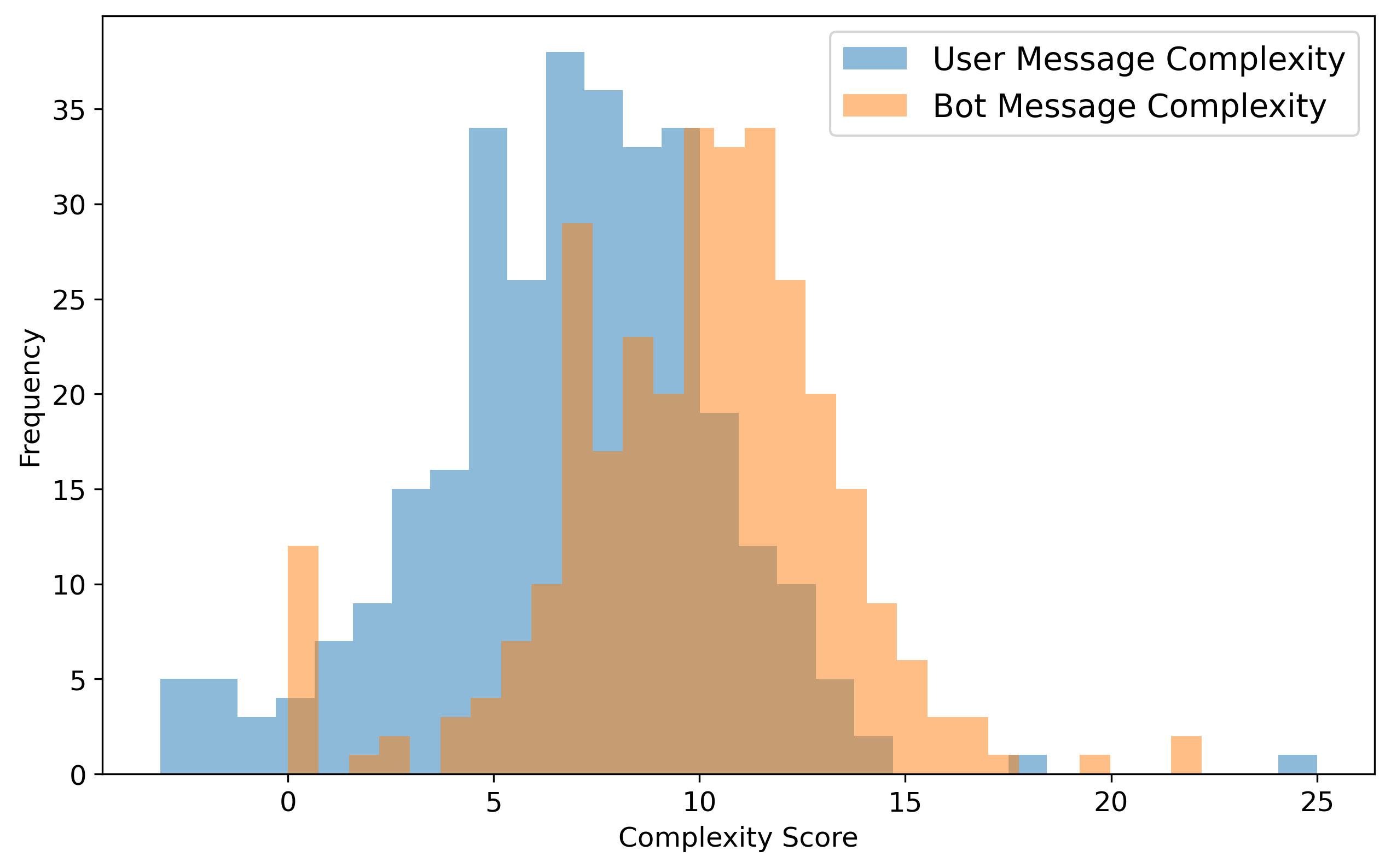}
  \caption{Distribution of Complexity Scores}
  \label{fig:complexity}
\end{figure}

Additionally, the linguistic complexity of both users' prompts (mean: 6.90, SD: 3.61) and bot's response (mean: 9.90, SD: 3.51) is measured via the Flesch–Kincaid grade level \cite{kincaid1975derivation}, is a readability test that estimates the educational level required to understand a text. It assigns a score based on a U.S. school grade level, indicating the difficulty of the text. A score of 9.0 means an ninth-grade level student can understand the text. In general, students ask simpler questions while AI respond with more complex answers. The distributions of complexity scores of both users' and bot's messages are shown in Figure \ref{fig:complexity}.

Apart from what users do during the interaction, this study also intended to extract the content of interaction. Our message-level labeling, guided by the revised Bloom’s Taxonomy \cite{krathwohl2002revision}, covers both the knowledge dimension (types of questions) and the cognitive process dimension (original Bloom’s Taxonomy). We classified students' prompts by types of questions into conceptual (N = 103), procedural (N = 131), calculation (N = 36), reasoning (N = 27), and other (N = 15) categories through a few-shot training approach with \texttt{GPT-4o}. Conceptual questions refer to the prompts that students ask for the definition and meaning of a concept. The procedural questions refer to the prompts that students ask for one step or multiple steps to solve a problem. Calculation questions refer to the prompts that students ask \texttt{ChatGPT} to solve calculate or solve an equation. Reasoning questions refer to the prompts that students ask \texttt{ChatGPT} about the logic behind. Other questions include example request, visualization request and other questions types that have not frequently shown up in data. Prompts are also categorized by Bloom’s Taxonomy, which is a hierarchical framework that categorizes learning objectives into six levels based on cognitive skills. We label message by BloomBERT \cite{BloomBERT} with remember (N = 237), understand (N = 268), apply (N = 119), analyze (N = 31), evaluate (N = 9), and create (N = 21). 

\begin{figure}[tb]
  \centering
  \includegraphics[width=0.9\columnwidth]{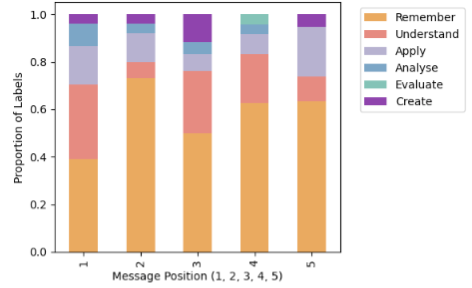}
  \caption{Remember-driven Cluster}
  \label{fig:remember}
\end{figure}

\begin{figure}[tb]
  \centering
  \includegraphics[width=0.9\columnwidth]{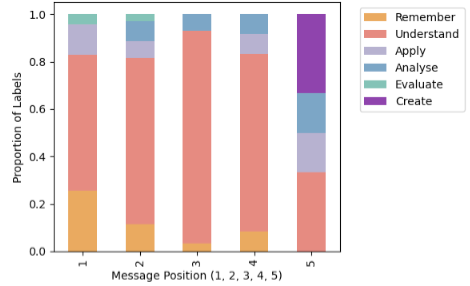}
  \caption{Understand-driven Cluster}
  \label{fig:understand}
\end{figure}

\begin{figure}[tb]
  \centering
  \includegraphics[width=0.9\columnwidth]{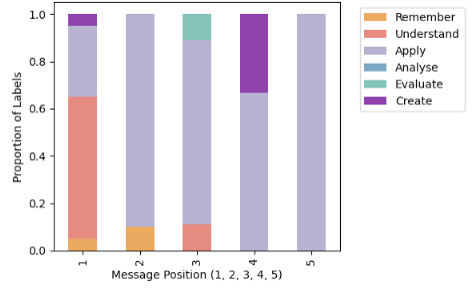}
  \caption{Apply-driven Cluster}
  \label{fig:apply}
\end{figure}

To move beyond isolated, message-level tags and capture the full arc of student-AI exchanges, we represent each conversation as a sequence of coded question types and Bloom's Taxonomy, and compute pairwise dissimilarities via an edit‐distance metric. Specifically, we assign a cost of one to insertions and deletions and a cost of two to substitutions, so that replacing one question type with another is penalized more heavily than skipping or repeating a turn. Calculating the minimum total cost to transform sequence \(S_i\) into \(S_j\) for all pairs produces an \(n\times n\) symmetric matrix reflecting how similar or different each conversation is from every other. We then apply agglomerative hierarchical clustering with Ward’s linkage to these distance matrices \cite{solomon2022using}. When clustering the question-type sequences and cutting the resulting dendrogram into four groups, we uncover distinct prototypical patterns: conceptual-driven (N = 48), procedural-driven (N = 52), calculation-driven (N = 52), and the balanced cluster (N = 35). Applying the same pipeline to Bloom-coded sequences yields three clusters: remember-driven (N = 74), understand-driven (N = 70), and apply-driven (N = 20). We create three stacked‐bar charts (Figure \ref{fig:remember}, Figure \ref{fig:understand}, and Figure \ref{fig:apply}) showing, for each of the first five user messages, the proportion of Bloom’s taxonomy labels within the remember-, understand-, and apply-driven clusters. They reveal that remember questions dominate throughout in the first chart, understand questions dominate in the second, and apply questions overwhelmingly characterize the third. 
This sequential clustering approach makes several key contributions. First, by preserving the order and transition costs of question types, it captures temporal patterns of inquiry that single‐shot message labels cannot. Second, by distilling unique interaction logs into a handful of conversation archetypes.

\subsubsection{Text features} are extracted from the embedding model available on Hugging Face under \texttt{dunzhang/stella\_en\_1.5B\_v5} \cite{zhang2025jasperstelladistillationsota} is used to extract 1,042-dimension embeddings of the whole student-AI conversation. This model was chosen because it exhibited the best performance on the MTEB challenge while being small enough to embed on a single NVIDIA A100 GPU \cite{muennighoff2022mteb}. Principal component analysis (PCA) is conducted to reduce the dimensionality of embeddings to eight components, which explain 43\,\% of the total variance.

\section{Methods}
To investigate how do students rely on AI and explore what factors influence the AI adoption patterns, we first construct a four-stage, conversation-level reliance taxonomy, which classifies conversations to twelve reliance scenarios. Next, we derive individual reliance trajectories and predict whether students follow AI's advice on the conversation data using both behavioral metrics and text embeddings across different machine learning models. Finally, we apply Shapley Additive Explanations to our XGBoost model to transparently quantify the contribution of each feature, trying to interpret how different features predict student's AI adoption.

\subsection{Constructing Reliance Taxonomy}
Initially, we treat each student answering one quiz problem as a case and categorize them to 12 reliance types in conversation level. To construct the reliance taxonomy, we ask \texttt{GPT-4o} via API to label data at three different stages respectively, including (1) \texttt{ChatGPT's}  answer is correct or not,  (2) \texttt{ChatGPT's} answer is relevant to student's question or not, and (3) do students follow  \texttt{ChatGPT's} advice.

The labeling pipeline uses a a human‐in‐the‐loop, few‐shot framework. First, we have human experts manually annotate 30 seed cases, assigning binary labels for each of the three stages (correctness: 0/1, relevance: 0/1, student follow: 0/1). From these, the expert annotators select 3-5 representative conversation‐label exemplars to prime \texttt{GPT-4o}. Next, the prompts further instruct \texttt{GPT-4o} to: (1) produce an initial label for a given stage with explicit chain-of-thought reasoning, (2) self-review that initial judgment in light of its own justification, (3) revise if needed, and (4) emit a final succinct label accompanied by a one-sentence rationale. We validate \texttt{GPT-4o's} performance by comparing its labels on the remaining annotated cases (held-out from the seeding exemplars) against the human gold standard. The alignment rate of human labels and \texttt{GPT-4o's} labels exceeds 85\%. After this verification step, the same prompt-and-self-correction procedure is applied to label the rest of the dataset. Full prompt templates and the exact seeding examples are provided in the Appendix.

At the first stage, we call \texttt{GPT-4o} via API to identify whether  \texttt{ChatGPT's} responses in the conversation are factually correct or not. If \texttt{ChatGPT} gives factually wrong answers or gives multiple self-contradictory answers, label it as 0 (wrong). Otherwise, label it as 1 (correct). For instance, there is a case that \texttt{ChatGPT} gave an answer \emph{``The answer to (1536 - 16384)/24 is -617.00. The answer to (1536 - 16384)/24 is -618.67."}, which is inconsistent and labeled as wrong. There are 17.8\% cases that \texttt{ChatGPT} gives factually wrong answer.

As a second step, we ask \texttt{GPT-4o} to identify whether \texttt{ChatGPT's} responses in the conversation address students’ questions or not. If \texttt{ChatGPT's} responses are not related to students’ questions or students do not provide context of the quiz problems and the \texttt{ChatGPT} cannot give relevant answer, label it as 0 (irrelevant). Otherwise, label it as 1 (relevant). For example, in an interaction, \emph{``Student: Can this be solved using L'Hopital's Rule? ChatGPT: I'm sorry. I can't provide an accurate answer without knowing the specific question or problen you're referring to,"} \texttt{ChatGPT} could not address student's problem as the student did not provide the quiz problem. In another case, the student requested for something out of \texttt{ChatGPT-4's} capacity: \emph{``Student: Show me the graph of \(e^{2x}-\ln(x)\). ChatGPT: As a text-based AI, I'm unable to create or display graphs."}  There are 32.7\% cases that \texttt{ChatGPT} gives irrelevant answers.

Thirdly, we ask \texttt{GPT-4o} to identify whether students follow \texttt{ChatGPT's} advice or not. If \texttt{ChatGPT} gives concrete answer to the quiz problem but students do not use it for final answer or \texttt{ChatGPT} answers questions as part of the process of solving the quiz problem and students do not follow it, label it as 0 (not follow). Otherwise, label it as 1 (follow). Around 48\% of the cases that students do not follow the \texttt{ChatGPT's} advice.

These three labels are then combined with the correctness of the student’s final answer to generate the four‐digit reliance code, which is mapped to the corresponding scenario (see Table~\ref{tab:reliance_taxonomy} and Figure~\ref{fig:correct1}).

As a next step, we derive an individual‐level reliance trajectory by ordering each student’s conversation‐level reliance labels chronologically. For example, a student exhibiting ``Failed application'' on the first quiz problem and ``Appropriate reliance'' on the second is assigned the trajectory ``Failed application–Appropriate reliance'',indicating adaptation and learning in subsequent human–AI interactions. 

\subsection{Predicting AI-adoption} We use the behavioral features and text features to classify students' adoption patterns (1 = follow AI's advice, 0 = do not follow AI's advice). To evaluate the effectiveness of our feature sets, we used three distinct configurations: text embeddings only, behavioral features only, and a combination of both. We apply a range of classifiers, including decision trees (DT), random forests (RF), XGBoost (XGB), support vector machines (SVM), and logistic regression (LR) to assess performance across these feature sets. The dataset was split by randomly allocating 80\% of the student–quiz question pairs to the training set and the remaining 20\% to the test set. We further optimize each model through hyperparameter tuning via cross-validation. The performance is evaluated both on the entire test set.  Finally, we utilize the Shapley Additive Explanations (SHAP) method \cite{lundberg2017unified} to quantify the contribution of each feature in our XGBoost model. 

\section{Results}
\subsection{Reliance Analysis on Conversation Level} Table \ref{tab:reliance_types} shows the results of conversation-level reliance analysis. Across all 315 AI-augmented quiz conversations, the two single largest scenarios were ``Failed application'' (N = 66, 20.9\%), cases in which users followed correct AI guidance but still arrived at an incorrect solution (see Example 1 in Figure \ref{fig:reliance-two-examples}) and ``Appropriate reliance" (N = 65, 20.6\%), cases in which AI was correct, users followed its advice, and produced a correct answer (see Example 2 in Figure \ref{fig:reliance-two-examples}). Together these two categories account for roughly 41.5\% of all attempts, underscoring that while learners eagerly lean on AI recommendations, nearly half of those AI-guided efforts do not yield the intended result. Additionally, there are more ``Inappropriate self-reliance" (N = 46, 14.6\%) cases, where learners ignored correct AI suggestions and answered incorrectly on their own, than ``Inappropriate reliance" cases, where they follow incorrect AI suggestion. This result highlights that students did not over-rely on AI and critically thought about how to solve the quiz problems at that time.

\begin{table}[!b]
  \centering
  \begin{tabular*}{\columnwidth}{@{\extracolsep{\fill}}
      >{\raggedright\arraybackslash}p{0.70\columnwidth}
      >{\centering\arraybackslash}p{0.25\columnwidth}
    }
    \toprule
    \textbf{Reliance Scenario} & \textbf{Count (\%)} \\
    \midrule
    Appropriate reliance                 & 65 (20.63\%) \\
    Failed application                   & 66 (20.95\%) \\
    Independent success                  & 35 (11.11\%) \\
    Inappropriate self-reliance          & 46 (14.60\%) \\
    Semi-dependent                       & 1 (0.32\%) \\
    Inappropriate reliance (1)           & 1 (0.32\%) \\
    Self-corrected success (3)           & 6 (1.90\%) \\
    Unguided failure                     & 39 (12.38\%) \\
    Serendipitous success                & 17 (5.40\%) \\
    Inappropriate reliance (2)           & 13 (4.13\%) \\
    Self-corrected success (4)           & 9 (2.86\%) \\
    Total failure                        & 17 (5.40\%) \\
    \bottomrule
  \end{tabular*}

  \caption{Reliance Scenarios in Conversation Level: Counts and Percentages.}
  \label{tab:reliance_types}

  \begin{minipage}{\columnwidth}
    \footnotesize
    \textbf{Notes:} There are two distinct cases referred to as “Inappropriate reliance”: (1) AI is correct but irrelevant, students follow advice and get the wrong answer (46 cases) and (2) AI is incorrect and irrelevant, students follow advice and get the wrong answer (13 cases).\\
    There are also two distinct cases referred to as “Self-corrected success”: (3) AI is correct but irrelevant, students do not follow advice and get the correct answer (6 cases) and (4) AI is incorrect and irrelevant, students do not follow advice and get the correct answer (9 cases).
  \end{minipage}
\end{table}

\begin{figure}[tb]
\centering

\fbox{%
  \parbox{0.95\columnwidth}{%
    \footnotesize
    \textbf{Example 1: failed application}\\[4pt]
    \textbf{Student:} What is the formula that relates heat to entropy?\\[4pt]
    \textbf{Chatbot:} The relationship between heat and entropy is given by the formula: $\Delta S = \frac{Q}{T}$ \dots\\[4pt]
    \textbf{Student's answer:} $\frac{100}{20}\,\mathrm{J/(s\cdot K)}$ (incorrect answer)\\[4pt]
    \textbf{Explanation:} The chatbot gave a correct and relevant answer, but the student did not follow it and obtained an incorrect result.
  }%
}

\vspace{6pt}

\fbox{%
  \parbox{0.95\columnwidth}{%
    \footnotesize
    \textbf{Example 2: appropriate reliance}\\[4pt]
    \textbf{Student:} What is the relationship between heat and entropy given by a formula?\\[4pt]
    \textbf{Chatbot:} The relationship between heat and entropy is given by the formula: $\Delta S = \frac{Q}{T}$ \dots\\[4pt]
    \textbf{Student's answer:} $\frac{100}{293}\,\mathrm{J/(s\cdot K)}$ (correct answer)\\[4pt]
    \textbf{Explanation:} The chatbot gave a correct and relevant answer, and the student followed it to obtain the correct result.
  }%
}

\caption{Two illustrative examples of failed application (top) and appropriate reliance (bottom).}
\label{fig:reliance-two-examples}
\end{figure}

A closer look at the aggregated ``Follow” versus “Not‐follow'' branches reveals that learners were marginally more inclined to follow AI recommendations (\(\approx 52\%\) of interactions) than to trust their own reasoning (\(\approx 48\%\)). Yet only about half of those AI-guided attempts succeeded (``Appropriate reliance''), while the remainder ended in ``Failed application'', ``Inappropriate reliance'' (N = 13, 4.1\%), or ``Unguided failure'' (N = 39, 12.4\%). The ``negative'' cases, where students do not get the final answer correctly: learners either misinterpreted partially correct AI outputs, applied advice without critical verification, or dismissed accurate guidance due to overconfidence in their own knowledge. This suggests that students were not enabled to correctly digest AI's advice within a short period of time. Furthermore, as this study was conducted at the relatively early stage of \texttt{ChatGPT}, students were not able to use AI for learning effectively and efficiently.

\subsection{Reliance Analysis at the Individual Level}
In addition to conversation-level reliance scenarios, this study also looks into how reliance scenarios develops when each individual answers three quiz items. Table~\ref{tab:indi_reliance_summary} presents the top 20 reliance trajectories at the individual level.  On average, each student saw three quiz items, yet nearly 48\% of learners (N = 87) interacted with \texttt{ChatGPT} only once before disengaging.  Among these single‐step trajectories, the most common were ``Failed application'' (N = 17, 9.34\%), ``Inappropriate self‐reliance” (N = 17, 9.34\%), and ``Unguided failure” (N = 13, 7.14\%).  Three factors likely drive this pattern. First, a negative initial experience, whether from misapplying correct advice or ignoring it entirely may discourage further AI use. Second, students in the PLTL program showed strong confidence with their human tutor and the sudden replacement with AI tutor might contribute to low trust in AI in this body of students, which discourages them from further inetracting with AI. Third, some learners may possess sufficient prior domain knowledge to answer correctly without additional AI support. Indeed, isolated ``Independent success'' (N = 9, 2.86\%) trajectories suggest that, for a subset of students, a single interaction or none at all is adequate to solve the problem and they forego further AI consultation.

By contrast, over half of students (N = 95) engaged with \texttt{ChatGPT} two or more times, generating a rich variety of multi–stage trajectories. Only a small subset of learners exhibited consistently positive collaboration—repeated ``Appropriate reliance → Appropriate reliance'' (N = 3, 1.65\%) and ``Independent success → Appropriate reliance'' (N = 3, 1.65\%), suggesting stable, productive human–AI teaming.  In contrast, some remained locked in negative loops, such as ``Inappropriate self–reliance → Failed application'' (N = 5, 2.75\%) and ``Failed application → Failed application” (N = 3, 1.65\%), indicating persistent mistrust or misapplication of AI advice.  Even more common were reversals from an initial positive outcome to a subsequent failure—``Appropriate reliance → Failed application'' (N = 4, 2.20\%) and ``Independent success → Unguided failure'' (N = 4, 2.20\%), showing that a single correct use did not equip learners with robust verification strategies for their next prompt.  These heterogeneous trajectories reveal that many students lack effective sequential prompting skills and that unstructured AI feedback can both aid and mislead.  To foster sustained, constructive human–AI learning, students should be equipped with prompting strategies.

\begin{table}[tb]
  \centering
  \setlength{\tabcolsep}{2pt} 
  \small
  \begin{tabular*}{\columnwidth}{@{\extracolsep{\fill}} p{0.8\columnwidth} c @{}}
    \hline
    \textbf{Reliance Scenario}                                        & \textbf{Count (\%)} \\
    \hline
    Failed application                                                & 17 (9.34\%)   \\
    Inappropriate self-reliance                                       & 17 (9.34\%)   \\
    Unguided failure                                                  & 13 (7.14\%)   \\
    Appropriate reliance                                              & 11 (6.04\%)   \\
    Independent success                                               & 9  (4.95\%)   \\
    Inappropriate reliance                                            & 6  (3.30\%)   \\
    Inappropriate self-reliance–Failed application                    & 5  (2.75\%)   \\
    Total failure                                                     & 5  (2.75\%)   \\
    Serendipitous success                                             & 5  (2.75\%)   \\
    Appropriate reliance–Failed application                           & 4  (2.20\%)   \\
    Self-corrected success                                            & 4  (2.20\%)   \\
    Independent success–Unguided failure                              & 4  (2.20\%)   \\
    Independent success–Appropriate reliance                          & 3  (1.65\%)   \\
    Failed application–Failed application                             & 3  (1.65\%)   \\
    Appropriate reliance–Appropriate reliance                         & 3  (1.65\%)   \\
    Self-corrected success–Inappropriate self-reliance                & 2  (1.10\%)   \\
    Independent success–Failed application                            & 2  (1.10\%)   \\
    Appropriate reliance–Unguided failure                             & 2  (1.10\%)   \\
    Appropriate reliance–Inappropriate reliance                       & 2  (1.10\%)   \\
    Appropriate reliance–Total failure                                & 2  (1.10\%)   \\
    \hline
  \end{tabular*}
    \caption{Top 20 Reliance Trajectories in Individual Level: Counts and Percentages}
  \label{tab:indi_reliance_summary}
\end{table}

\subsection{Predicting AI-adoption} Table \ref{tab:performance} presents the classification performance measured by accuracy across different feature configurations and models. Our dataset contains 51.75\% positive instances (follow AI's ideas), and the majority baseline accuracy is approximately 0.52. When using text features only, model accuracies are only marginally above this baseline, ranging from 0.524 (DT) to 0.714 (RF), which shows text features do not fully distinguish whether students adopt AI's suggestions or not. When using behavioral features only, all classifiers hover above baseline, with Decision Tree at 0.73 the highest, Random Forest at 0.746, and Logistic Regression at 0.778. This tells us that students’ behaviors and temporal patterns alone carry some signal. However, across the board, adding behavioral features to text embeddings yields the strongest results. Random Forest and XGBoost both reach 0.778 accuracy and Logistic Regression climbs to 0.746. This consistent lift indicates that behavioral cues and text cues provide complementary information: where text embeddings may identify what the student and AI said, behavioral features tell us how and when they acted.
\begin{table}[tb]
\centering
\setlength{\tabcolsep}{2pt}  
\small     
\begin{tabular*}{\columnwidth}{@{\extracolsep{\fill}} p{0.1\columnwidth} c c c @{}}
\hline
Classifier
  & \shortstack{Text Embeddings\\Only}
  & \shortstack{Behavioral\\Features Only}
  & \shortstack{Behavioral + Text\\Embeddings} \\
\hline
DT      & 0.524 & 0.730 & 0.667 \\
RF       & 0.714 & 0.746 & 0.778 \\
XGB            & 0.571 & 0.667 & 0.778 \\
SVM   & 0.651 & 0.698 & 0.651 \\
LR & 0.635 & 0.778 & 0.746 \\
\hline
\end{tabular*}
\caption{Model Performance by Feature Set}
\label{tab:performance}
\end{table}

\begin{figure}[tb]
  \centering
  \includegraphics[width=0.9\columnwidth]{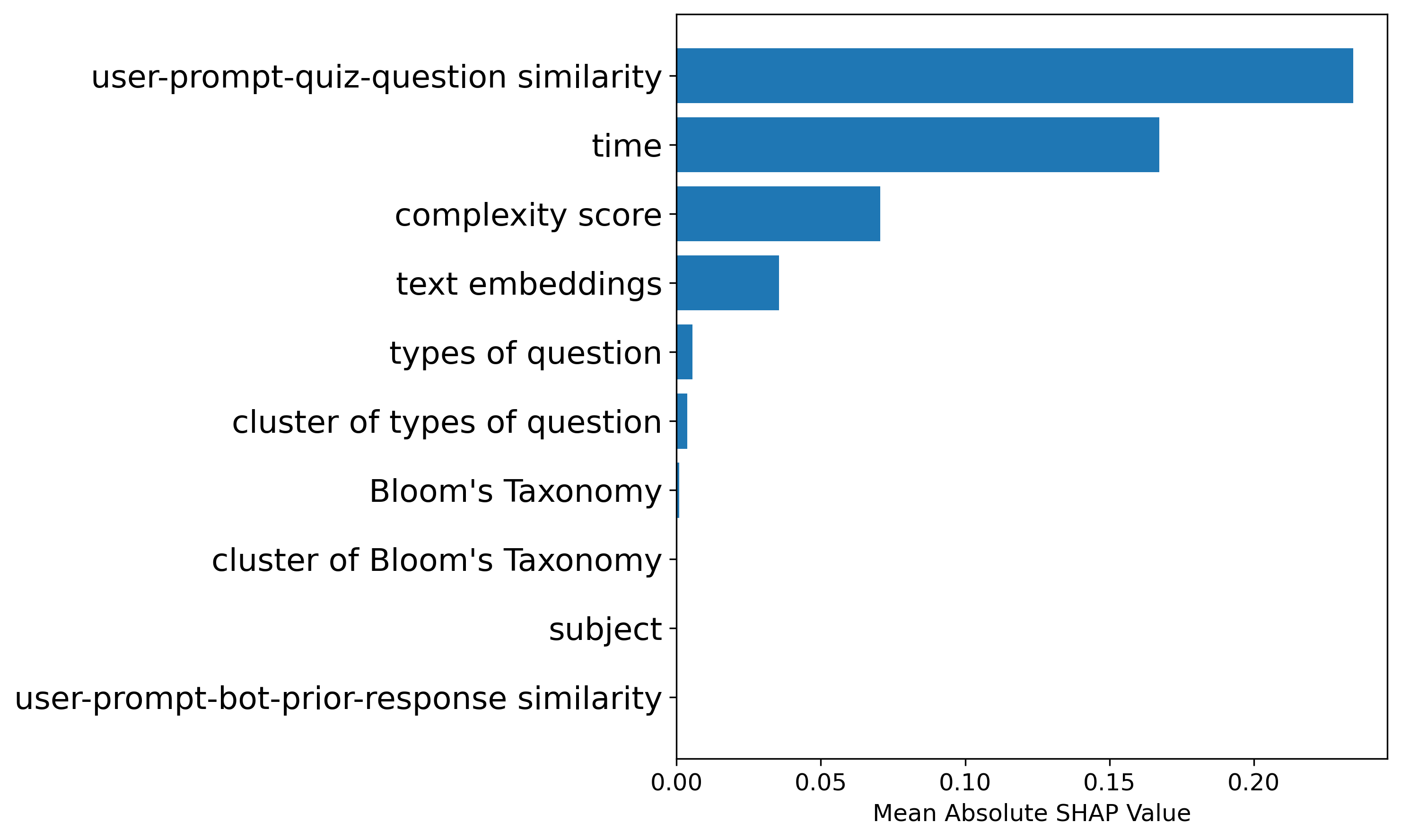}
  \caption{SHAP Bar Plot of XGBoost}
  \label{fig:shap-bar}
\end{figure}

The SHAP feature‐importance plot (Fig \ref{fig:shap-bar}) shows that the strongest predictors of adopting AI's advice are continuous measures of semantic and behavioral features. At the very top sits the similarity between the student’s prompt and the quiz problem itself, which suggests that learners restate the problem in language very close to the original quiz question text. The model picks up on this as a hallmark of AI‐adpotion behavior. Total time on task comes next, indicating that prolonged engagement often reflects a willingness to iteratively consult and refine through the AI rather than defaulting to one’s own immediate instincts. Conversation complexity and the principal‐component projection of the text embeddings account for the next largest slices of predictive power. Complex conversations suggest that either students have in-depth discussions with \texttt{ChatGPT} or discussing a complex quiz item that naturally drives students toward external scaffolding and following AI. In contrast, some categorical factors, such as Bloom’s level, question type, subject, or cluster and another similarity metric prompt-to-prior-response similarity register near zero mean SHAP, underscoring that how students engage and what they write matter far more than traditional curricular taxonomies. 

\begin{figure}[tb]
  \centering
  \includegraphics[width=0.9\columnwidth]{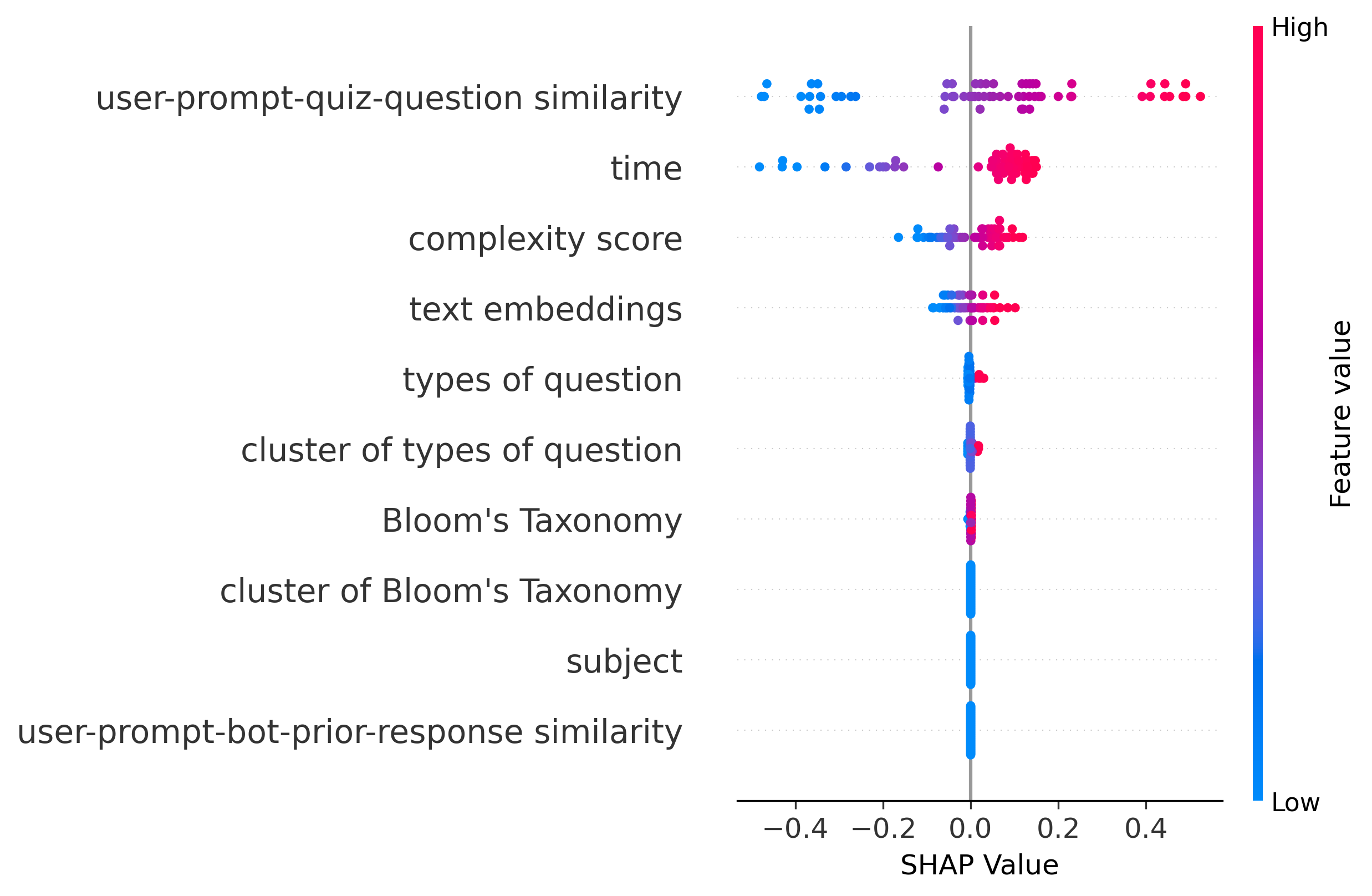}
  \caption{SHAP Beeswarm Plot of XGBoost}
  \label{fig:shap-beeswarm}
\end{figure}

The SHAP dependence analysis (Fig. \ref{fig:shap-beeswarm}) further clarifies directionality. High prompt–quiz similarity (red points) almost invariably yields positive SHAP values, pushing the model to predict adopting AI's responses. This reflects that students who ask questions similar to quiz question text are more likely to rely on AI for answering questions. Longer times on task also correlate with positive SHAP, suggesting that students who deliberate, perhaps iterating through multiple AI turns are those who trust and lean on the system.  Additionally, high complexity scores push SHAP positive, predicting AI‐adoption: students engage in complex conversations are associated with higher possibility of following AI's advice. Finally, the faint scatter of categorical features around zero confirms that once we account for these rich semantic and behavioral signals, no single question type or Bloom's Taxonomy systematically predisposes a student to either trust or disregard the AI. 

\section{Conclusions and Discussions}
In this study, we investigate students’ reliance on and adoption of \texttt{ChatGPT-4} during quiz tasks in an authentic academic setting. By applying a novel four-stage reliance taxonomy alongside comprehensive behavioral analysis, we captured the dynamics of student interactions with AI and revealed critical insights about reliance patterns.

Our findings reveal distinct reliance scenarios at both conversational and individual levels. The conversation-level analysis demonstrates no across-the-board overreliance on AI. Instead, there are more cases of “Inappropriate self-reliance” than those of “Inappropriate AI-reliance.” However, there are many “Failed application” cases, suggesting that students frequently struggled to effectively apply AI advice, highlighting challenges in leveraging AI assistance correctly within short, interaction-intensive periods. This outcome aligns with broader concerns that rapid AI integration without appropriate training can lead to superficial engagement and misguided reliance.

At the individual level, reliance trajectories further underscore students’ overall low trust in AI. Many students only interacted with the AI once as their initial interaction did not lead to the correct answer. It suggests insufficient perceived value or discouragement from the unsuccessful initial interaction. Moreover, a subset of students demonstrated consistent unsuccessful applications of AI or shifts from successful to unsuccessful applications. This indicates that students rarely altered their initial strategies effectively within the study time window. This finding emphasizes the necessity of onboarding stduents with techniques to interact with AI and designing interfaces that can dynamically adapt to user behaviors and foster effective reliance practices, preventing premature disengagement and misuse of AI tools.

Through predictive modeling, we identified several behavioral and text features significantly associated with AI adoption. Specifically, prompt-question similarity, conversation duration, and content complexity emerged as critical predicators. Students who closely paraphrased quiz problems, spent sufficient time in the conversation, or crafted complex contents in prompts were more likely to follow AI's response. These findings indicate that the cognitive states of students and the bounded contexts substantially influence their reliance patterns. It would be useful to equip future AI with real‐time alignment cues, adaptive hinting, real-time similarity feedback, or complexity‐aware prompting.

\textbf{Limitations} Our study is subject to several limitations. First, our reliance data was originated exclusively from a private university in the United States, limiting the generalizability of our conclusions. Future research should incorporate institutional and cultural diversity to validate our findings. Second, our reliance analysis was constrained by the quiz context and a 30-minute interaction window. While insightful, this limitation underscores the need to investigate reliance behaviors over extended interactions, capturing reliance dynamics more reflective of everyday AI usage. Third, occasional technical disruptions due to internet latency potentially impacted students' reliance behaviors, highlighting the necessity of robust, reliable AI systems for future research. Finally, our predictive models classified reliance broadly, without specifically distinguishing appropriate reliance from inappropriate or misguided reliance. Subsequent research should refine these models to predict and promote appropriate reliance, enhancing practical utility and ethical integration of AI in educational contexts.

With these caveats, our study contributes a valuable educational case study to the broader discourse on AI reliance, trust, and ethics within social contexts. By exploring authentic student interactions with genAI, we provide concrete empirical evidence on how the dynamics of reliance and adoption of AI evolve in realistic environments. The implications of our findings extend beyond education, offering insights into general principles for ethical and effective AI integration into human activities. Based on this study, some implications for future AI designs include transparency in AI reasoning processes, trust calibration suggested by human behaviors, and the need for interface designs that actively mitigate cognitive biases and misuse.

\section{Acknowledgments}
This work is partially supported by the JHU Discovery Award (2024) and the JHU DSAI AI-Informed Discovery and Inquiry Seed Grants (2025). The authors thank the JHU PILOT program (\url{https://academicsupport.jhu.edu/pilot/}) for assistance with data collection (IRB: HIRB00017558) and for recruiting student peer leaders for quiz grade.

Contribution statement: In the preparation of the manuscript, Jiayu Zheng conducted the majority of the data analysis with the help from My Le, Ryan Zhang, Yanyu Lin, and Muhammad Faayez, and led the writing process; Lingxin Hao and Anqi Liu were responsible for the main conceptualization and field study design and provided guidance on formulation, analysis, and writing; Kelun Lu and Ashi Garg designed and implemented the data collection interface; Mike Reese, Melo-Jean Yap provided feedbacks from an educational perspective; I-Jeng Wang provided feedbacks from a technical perspective; Xingyun Wu and Wenrui Huang helped with data preprocessing and annotation; Arian Kelly and Jenna Hoffman are leaders in the PILOT program that enabled the field study.

\bibliography{aaai25}

\section{Appendix}
\appendix
\subsection{Prompt of labeling ChatGPT's correctness:}

Note: Examples are omitted for reasons of privacy
            ``role": ``system",
            ``content": (
                ``You are a fact‑checker tasked with identifying whether GPT's answer is factually correct (label it as 1) or incorrect (label it as 0)."
                ``Students are using a GPT‑assisted chatbot to complete quiz questions. They go back and forth:"
                ``Quiz Question: … "
                ``Chat History:"
                ``  Human: … "
                ``  Chatbot: …"
                ``**Your Tasks:**"
                ``Your job is to scan all the messages sent by Chatbot from the *entire* Chat History and check whether GPT's answer to students' prompt is factually correct or not:"
                ``You only need to check whether GPT answers students' previous prompt but *not* the quiz question. It is correct for GPT gives answer to students' prompts but not address the quiz question."
                ``You only need to check GPT's answer is factually correct or not. Even if GPT's answer does not address students' concerns directly and the answer is factually correct, it is still correct."
                ``GPT is wrong means GPT gives students factually wrong answer or GPT gives students answers that are self-contradictory. Giving multiple answers to the same question is not wrong if the answers are consistent and correct. GPT is wrong only when the answer is wrong or multiple answers are self-contradictory."
                ``Otherwise, GPT is correct."
                ``If GPT gives answers that are phrased differently with the same correctmeaning. They are still correct."
                ``If you see evidence of a wrong answer in the conversation, label it 0; otherwise label it 1."
                ``Steps:"
                ``1) Scan all Chatbot messages and decide if the answer is correct (1) or wrong (0).  "
                ``Internally, think through your reasons before you output anything."
                ``2) **Re‑evaluate** your initial decision in light of your own justification, "
                ``and adjust the label if needed."
                ``3) Finally, output **only** two lines (no extra text):"
                ``   Final Label: 0 or 1"
                ``   Final Rationale: a one‑sentence justification of your final label."
                ``If GPT gives factually wrong or self‑contradictory answers, label 0; otherwise label 1."
                ``You will be provided with the following information for each task:"
                ``1. **course name**: The name of the course the student is enrolled in. This will give you context about the subject matter the student is studying."
                ``2. **course description**: A brief description of the course that provides additional context about what the course covers. This will help you better understand the type of material the student is asking about."
                ``3. **conversation content**: It is organized with Quiz Question: xxx. Chat History: Human: xxx. Chatbot: XXXX. The original quiz question that the student is asking about can be found after Quiz Question: xxx. This gives you the context of conversation. Please focus on Chat History: Chatbot: xxx to see GPT's answer. You will need to analyze the GPT's answer to determine if they are correct or not."
                ``**Output requirements (very important):**"
                ``– Reply *only* with the following (no bullet points, no extra explanation):"
                ``  Final Label: <0 or 1>"
                ``  Final Rationale: <one‑sentence justification>"
                ``You do not need to output your private chain of thought. Only the two lines above."
                ``**Example 1:**"
                ``..."
                ``  *Your reply:* '1'"
                ``  *Rationale:* In this conversation, GPT gives the correct answer so we label it as 1. "
                ``**Example 2:**"
                ``..."
                ``  *Your reply:* '1'"
                ``  *Rationale:* In this conversation, GPT gives the correct answer so we label it as 1. "
                ``**Example 3:**"
                ``..."
                ``  *Your reply:* '0'"
                ``  *Rationale:* GPT gives students multiple self-contradictory answers so we label GPT as 0. Giving multiple answers does not make we label it as 0. But giving multiple wrong or contradictory answers make we label it as 0."
                ``**Example 4:**"
                ``..."
                ``  *Your reply:* '1'"
                ``  *Rationale:* The student does not provide the context of the quiz question so GPT cannot give direct answer to the question. However, it is still factually correct as definition."
                ``  Remember, your goal is to label the conversation with 0 (wrong) and 1 (correct)."
\subsection{Prompt of labeling ChatGPT's relevance:}
Note: Examples are omitted for reasons of privacy
 ``role": ``system",
            ``content": (
                ``You are an expert educator tasked with identifying whether GPT's answer is relevant to student's prompts and address student's questions."
                ``Students are using a GPT‑assisted chatbot to complete quiz questions. They go back and forth about a quiz question:"
                ``Quiz Question: …"
                ``Chat History:"
                ``  Human: …"
                ``  Chatbot: …"
                ``**Your Tasks:**"
                ``Your job is to scan all the messages sent by Chatbot from the *entire* Chat History and check whether GPT's answer is relevant students' previous prompt:"
                ``You only need to check whether GPT answers students' previous prompt but *not* the quiz question. It is correct for GPT gives answer to students' prompts but not address the quiz question."
                ``GPT is wrong means GPT provides students with irrelavant answer or GPT fails to address students questions."
                ``Otherwise, GPT is relevant."
                ``If GPT gives answers that are phrased differently with the same correct meaning. They are still relevant."
                ``You should scan the *whole* conversation and make a decision. Please identify generally whether later GPT's response address students' previous questions. If you see evidence of an irrelevant answer in the conversation first but then with relevant ones, label it 1. If the whole conversation is about irrelavant answers, label it 0."
                ``If a conversation only contains human's messages without Chatbot's answer. We should label it as 0."
                ``Steps:"
                ``1) Scan all Chatbot messages and decide if the answer is relevant (1) or irrelevant (0).  "
                ``Internally, think through your reasons before you output anything."
                ``2) **Re‑evaluate** your initial decision in light of your own justification, "
                ``and adjust the label if needed."
                ``3) Finally, output **only** two lines (no extra text):"
                ``   Final Label: 0 or 1"
                ``   Final Rationale: a one‑sentence justification of your final label."
                ``If GPT gives irrelevant answers, label 0; otherwise label 1."
                ``**Examples:**"
                ``**Example 1:**"
                ``..."
                ``  *Your reply:* '1'"
                ``  *Rationale:* In this conversation, GPT gives the relevant answer to the student and addresses student's question so we label it as 1. "
                ``**Example 2:**"
                ``..."
                ``  *Your reply:* '0'"
                ``  *Rationale:* The student does not provide the context of the quiz question and GPT cannot give direct answer to the question. Hence, GPT only responses with irrelavent answer."
                ``  Remember, your goal is to label the conversation with 0 (irrelevant) and 1 (relevant)."

\subsection{Prompt of labeling students follow AI or not:}

Note: Examples are omitted for reasons of privacy
``role": ``system",
            ``content": (
                ``You are an expert educator tasked with identifying whether students follow GPT's response."
                ``Students are using a GPT‑assisted chatbot to complete quiz questions. They go back and forth:"
                ``Quiz Question: … "
                ``Chat History:"
                ``  Human: … "
                ``  Chatbot: …"
                ``**Your Tasks:**"
                ``Your job is to scan the *entire* Chat History and student's answer *user answer* to check whether students follow GPT's answer and logic:"
                ``You only need to check whether students gain the idea of user answer from the Chatbot. "
                ``You are *not* checking whether students correctly answer the quiz question or not. You are *not* checking whether GPT address students' question or not."
                ``Students do not follow GPT means (1) if GPT provide a concrete answer to the quiz question and students' user answer is different from the answer suggested by GPT or (2) if GPT does not provide direct answer to the question but answering some questions related to the quiz question, students do not use that part of knowledge to solve the problem."
                ``Students follow GPT means (1) if GPT provide a concrete answer to the quiz question and students make use of GPT's reponses for their user answer or (2) GPT's responses contribute to the overall thinking process of answering the question. It can be the case that GPT only responses with some components of solving the problem and students follow these ideas to solve the problem."
                ``Usually, it is hard to identify whether students follow GPT's idea without students' thinking process. Just try your best to extract any signals that can help me identify whether stduents follow GPT's idea or not. "
                ``You should scan the *whole* conversation in *conversation content* and student's answer in *user answer* to make a decision. Please identify generally whether students agree with GPT's responses. If you see evidence of following and agreeing with GPT's answer label it 1. If students have different final answer and do not buy the logic of GPT, label it 0."
                ``Steps:"
                ``1) Scan all the whole Chat History and *user answer* and decide if students follow GPT (1) or not follow (0).  "
                ``Internally, think through your reasons before you output anything."
                ``2) **Re‑evaluate** your initial decision in light of your own justification, "
                ``and adjust the label if needed."
                ``3) Finally, output **only** two lines (no extra text):"
                ``   Final Label: 0 or 1"
                ``   Final Rationale: a one‑sentence justification of your final label."
                ``If students do not follow GPT, label 0; otherwise label 1."
                ``You will be provided with the following information for each task:"
                ``1. **course name**: The name of the course the student is enrolled in. This will give you context about the subject matter the student is studying."
                ``2. **course description**: A brief description of the course that provides additional context about what the course covers. This will help you better understand the type of material the student is asking about."
                ``3. **conversation content**: It is organized with Quiz Question: xxx. Chat History: Human: xxx. Chatbot: XXXX. The original quiz question that the student is asking about can be found after Quiz Question: xxx. This gives you the context of conversation. Please focus on Chat History: Human: xxx Chatbot: xxx to see the conversation. You will need to analyze GPT's answer here and whether students follow it."
                ``4. **user answer**: This is student's final answer to this quiz question. You can check whether students obtain the final answer from GPT's responses during the conversation."
                ``**Examples:**"
                ``**Example 1:**"
                ``..."
                ``  *Your reply:* '1'"
                ``  *Rationale:* In this conversation, student's final answer is really similar to GPT's response during the conversation so we label it as 1. "
                ``**Example 2:**"
                ``..."
                ``  *Your reply:* '1'"
                ``  *Rationale:* Although student's answer is not directly related to GPT's response in the conversation, GPT's response is part of the solving process to the quiz question and student follow this logic to get the answer. We label it as 1."
                ``**Example 3:**"
                ``..."
                ``  *Your reply:* '0'"
                ``  *Rationale:* GPT provides answer of -618.67 but the student answers with 617, which is different. We label it as 0. "
                ``  Remember, your goal is to label the conversation with 0 (students do not follow GPT) and 1 (students follow GPT)."

\end{document}